\documentclass[letterpaper, 10 pt, conference]{ieeeconf}  % Comment this line out if you need a4paper

\IEEEoverridecommandlockouts                              % This command is only needed if 
\newcommand{\tabincell}[2]{\begin{tabular}{@{}#1@{}}#2\end{tabular}} 
   \usepackage[pdftex]{graphicx}
   \graphicspath{{./pic/}}
   \DeclareGraphicsExtensions{.pdf,.jpeg,.png}
\usepackage[justification=centering]{caption}
\usepackage{float}
\usepackage{cite}
\usepackage{textcomp}
\usepackage{multirow}
\usepackage{balance}
  \usepackage[caption=false,font=normalsize,labelfont=sf,textfont=sf]{subfig}
\title{\LARGE \bf
Characterization of a RS-LiDAR for 3D Perception
}

\author{Zhe Wang, Yang Liu, Qinghai Liao, Haoyang Ye, Ming Liu and Lujia Wang% <-this % stops a space
\thanks{Zhe Wang, Yang Liu, Qinghai Liao, Haoyang Ye and Ming Liu are with Robotics and Multi-Perception Lab (RAM-LAB), Robotics Institute, The Hong Kong University of Science and Technology}
\thanks{Lujia Wang is with Cloud Computing Lab of Shenzhen Institutes of Advanced Technology, Chinese Academy of Sciences}
}
\begin{document}

\maketitle
\thispagestyle{empty}
\pagestyle{empty}

%%%%%%%%%%%%%%%%%%%%%%%%%%%%%%%%%%%%%%%%%%%%%%%%%%%%%%%%%%%%%%%%%%%%%%%%%%%%%%%%
\begin{abstract}

High precision 3D LiDARs are still expensive and hard to acquire. This paper presents the characteristics of RS-LiDAR, a model of low-cost LiDAR with sufficient supplies, in comparison with VLP-16. The paper also provides a set of evaluations to analyze the characterizations and performances of LiDARs sensors. This work analyzes multiple properties, such as drift effects, distance effects, color effects and sensor orientation effects, in the context of 3D perception. By comparing with Velodyne LiDAR, we found RS-LiDAR as a cheaper and acquirable substitute of VLP-16 with similar efficiency.

\end{abstract}

%%%%%%%%%%%%%%%%%%%%%%%%%%%%%%%%%%%%%%%%%%%%%%%%%%%%%%%%%%%%%%%%%%%%%%%%%%%%%%%%
\section{Introduction}

\subsection{Motivation}
Along with the rapid development of autonomous driving, multi-beam LiDAR has become one of the most important sensors on autonomous cars. Light Detection and Ranging, known as LiDAR, is a system using lasers to mainly detect the geometrical properties such as location, shape, and velocity. It beams lasers to the target objects and receives the signals reflected by those objects. By comparing the phase difference between the received signals with the sent ones, it reveals information about the target objects, for example, the distance, reflectivity, etc. By adopting further algorithms, the position, altitude, velocity, pose, and shapecould also be obtained \cite{Kneip},\cite{Glennie}. Generally, LiDAR is capable of detecting targets with a precision of several centimeters. 

LiDAR on vehicles is the critical sensor that serves for mapping and localization, for which multiple products have been developed. Velodyne LiDAR is widely considered to be the most popular LiDAR company who has released several LiDAR products such as HDL-64E, HDL-32E and VLP-16 for mapping and 3D perception purposes \cite{Levinson}. VLP-16 has generated significant interest in the surveying and mapping industry because of its compact size, low power requirements, and high performance. However, Velodyne LiDAR are comparably expensive and buyers should wait for at least 6 months to get the sensor. At present, plenty of low-cost LiDAR products have come out and two of the most famous ones are RS-LiDAR from Robosense and PANDAR 40 from HESAI. 

\begin{figure}
\centering
\includegraphics[height=1in]{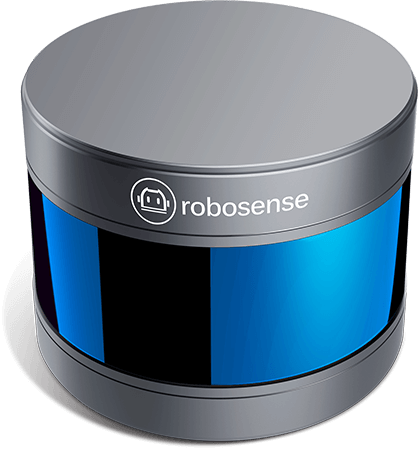}
\caption{RS-LiDAR from Robosense \protect\\ (Retail price: about \$7000)}
\label{fig_rslidar}
\end{figure}

For RS-LiDAR is also very popular in autonomous driving companies like TuSimple\footnote{http://www.tusimple.com}, and RoadStar\footnote{http://roadstar.ai}, we evaluate its characteristics and justify wether it  is feasible as replacement of the VLP-16. As the same as VLP-16, RS-LiDAR is a 16-channel real-time 3D LiDAR with a similar dimensions and weights. The RS-LiDAR is shown in Fig.\ref{fig_rslidar} is designed to be used on autonomous cars, robots, and UAVs. Usually, these applications require high accuracy and proper size and weight. With knowledge of the parameters in the official handbook, we consider it is still necessary to have a systematic evaluation of its accuracy, repeatability, and stability.It is worth to mention that there are some low-cost 2D LiDARs such as RpLiDar A2 and A1 from SlamTec\footnote{https://www.slamtec.com}, which can be assembled with rotational parts and work as 3D sensor. As the scanning frequency is much lower comparatively, we only consider the 3D LiDARs in this paper.

\subsection{Related Work}
For laser scanner evaluation, many papers had been published with different experiments for different sensors. Kneip proposed a characterization and extended some specialized tests with a subsequent calibration model of a 2D LiDAR, URG-04LX\cite{Kneip}. Glennie put forward a calibration and stability analysis of the VLP-16 laser scanner\cite{Glennie}. Ye and Borenstein presented a characterization study of the Sick LMS 200 laser scanner\cite{CangYe}. Stone reviewed the basic physics and implementation of various LADAR technologies, describing the problems associated with available 'off-the-shelf' LADAR systems and summarizing worldwide state-of-the-art research. He also elaborated on general trends in advanced LADAR sensor research and their likely impact on manufacturing, autonomous vehicle mobility and on construction automation\cite{WilliamCSton}. Kawata introduced a method to develop an ultra-small lightweight optical range sensor system\cite{Hirohiko} and Ueda proposed an accurate range data mapping system with sensor motion\cite{Tatsuro}. According to their tests and analyses, we detected the commonly concerned issues with RS-LiDAR and VLP-16.
\subsection{Contributions}
Our investigation in this paper is targeted to evaluate the RS-LiDAR's essential characteristics and make comparisons with VLP-16. The paper investigates the drift effects, influences of sensors orientation, target surface color and distance of RS-LiDAR. There were also tests on common scenes or objects and tests on the road with RS-LiDAR and VLP-16. These experiments and tests methods can be easily generalized to evaluate a LiDAR's performances with comparisons with others, up to the interest of readers.
\subsection{Organization}
The remainder of the paper is organized as: Section II is an introduction to RS-LiDAR and features comparison with VLP-16. Section III presents the details of the experiments, their results and analysis. Section IV investigates the calibration methods for RS-LiDAR, followed by conclusion in
Section V.

\section{The Labeled Characteristics}%--------------------------------------------------2222222222222222222222222222
\label{sec_feature}
The RS-LiDAR is a 16-channel solid-state hybrid LiDAR developed by Suteng Innovation Technology Co., Ltd. Its features are shown in TABLE \ref{features}, in comparison with VLP-16. Most of the labeled features of these two LiDARs are identical or very close. RS-LiDAR owns better features in the accuracy, measurement range, data points generated and price, while VLP-16 has the features in power consumption, weight, and dimensions with marginal advantages.
\begin{table}[b]
\renewcommand{\arraystretch}{1.3}
\caption{Features of VLP-16 and RS-LiDAR}\label{features}
\scriptsize
\centering
\begin{tabular}{l|c|c}
\hline
\bfseries Features & \bfseries VLP-16 &\bfseries RS-LiDAR\\
\hline
Channels & 16  & 16  \\
Wavelength & 903 nm & 905 nm \\
Laser Product Classification & class 1 & class 1 \\
Accuracy & $\pm$3 cm (Typical) & $\pm$2 cm (Typical) \\
Measurement Range & Up to 100m & 20cm $\sim$ 150m \\
Single Return Data Points & 300000 pts/s & 320000 pts/s \\
Field of View (Vertical) & $30^\circ$ & $30^\circ$ \\
Angular Resolution (Vertical) & $2.0^\circ$ & $2.0^\circ$\\
Field of View (Horizontal) & $360^\circ$ & $360^\circ$ \\
Horizontal Angular Resolution & $0.1^\circ-0.4^\circ$ & $0.1^\circ-0.4^\circ$\\
Rotation Rate & 5-20 Hz & 5-20 Hz\\
Power Consumption & 8 W (Typical) & 9 W (Typical)\\
Environmental Protection & IP67 & IP67\\
Operating Temperature &$-10\sim60^\circ$C& $-10\sim60^\circ$C\\
Weight & 830 g & 840 g\\
Dimensions&$\phi$ 103mm, H 72mm&$\phi$ 109mm, H 82.7mm\\
Retail Price&\$7999 & \$7000\\
\hline
\end{tabular}
\end{table}

\section{Evaluation of the RS-LiDAR}%--------------------------------------------------33333333333333333333333
\label{sec_experi}
This section presents the experiments to test characteristics of RS-LiDAR and VLP-16, as well as the results of the experiments. The section analyzes the performance of two LiDARs in the aspects of drift effects caused by the temperature, sensor orientation influence, differents of the 16 laser beams, surface color influence and different representations with different target distances.

\subsection{Drift Effects}
It is expected to have embedded correlation with temperature to revise the drift effect errors of a laser scanner. So in general, a majority of LiDAR instruments contain range corrections corresponding to the internal operating temperature of the laser and detector pairs. The drift effect reveals the stability of the LiDAR. 

\begin{figure}
\centering
\includegraphics[width=\columnwidth]{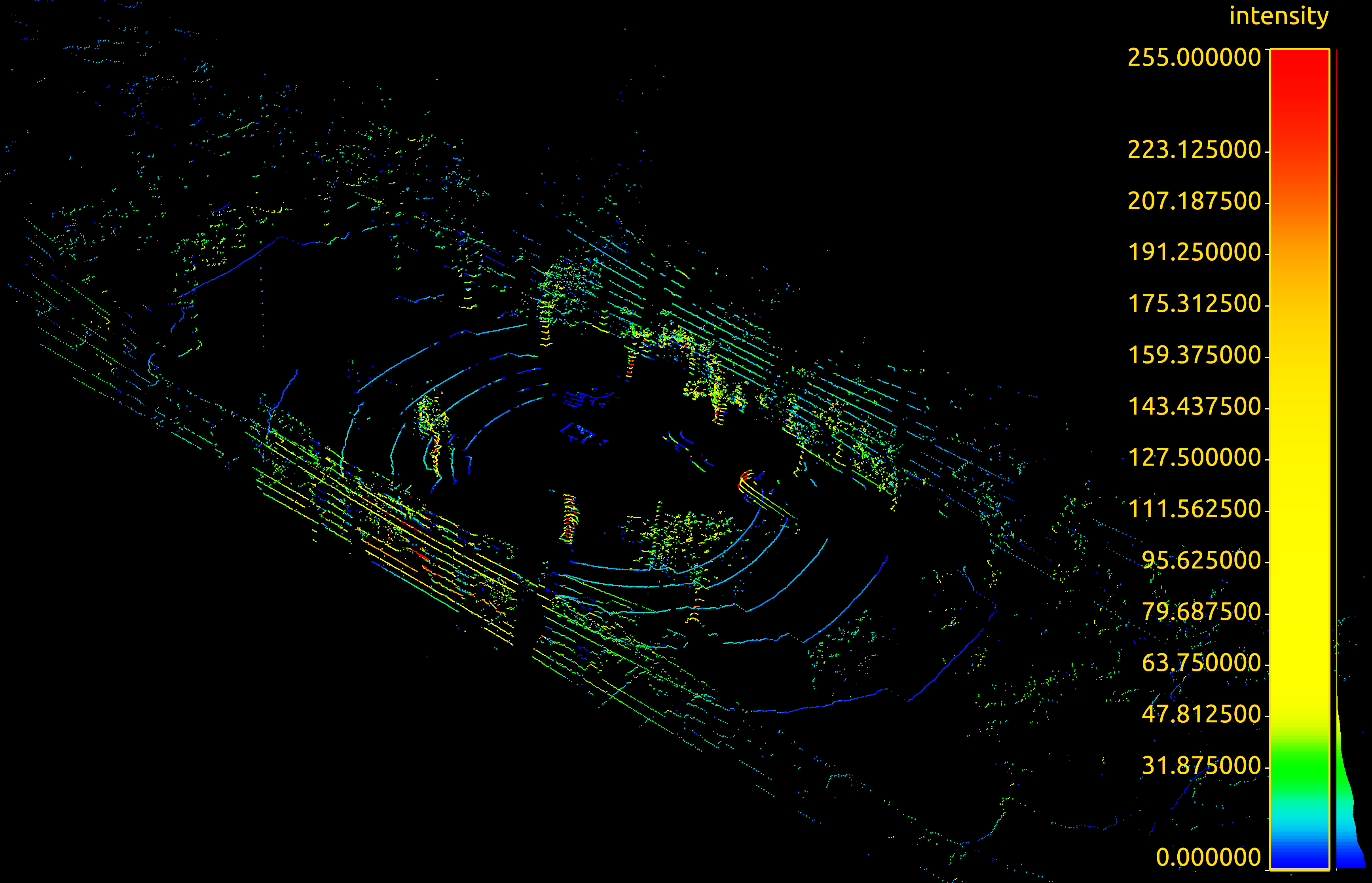}
%\DeclareGraphicsExtensions
\caption{Typical point-cloud collected by the RS-LiDAR}
\label{pointsurf}
\end{figure}

To get the drift effect and analyze the stability of RS-LiDAR, measurements of a plane surface with a long time has been performed. We settled RS-LiDAR at $1.45$ meters far away from a white wall and kept the y-axis of LiDAR right vertical to the plane of the wall\footnote{Note that we use motion tracking system to provide precise pose estimation to all the tests included in this paper.}. We are not able to get the temperature values of the LiDAR core, but the temperature will go up with the device keeps working. So we measured the distances of the wall with RS-LiDAR for several times and it is lasted for about $60$ minutes for each measurement. At each time, we started the test when the LiDAR had been cooled down to the ambient temperature of $27^\circ$C. The computer recorded a datum every $10$ frames with sensor working at about $0.2^\circ$ horizontal angular resolution. Fig.\ref{pointsurf} presents the data points collected by a LiDAR and the color of each point represents for its reflection intensity. In order to get the distance values from the collected data points, we segmented the target area and computed the mean value of these points' y-axis values. 

\begin{figure}
\centering
\includegraphics[width=3.3in]{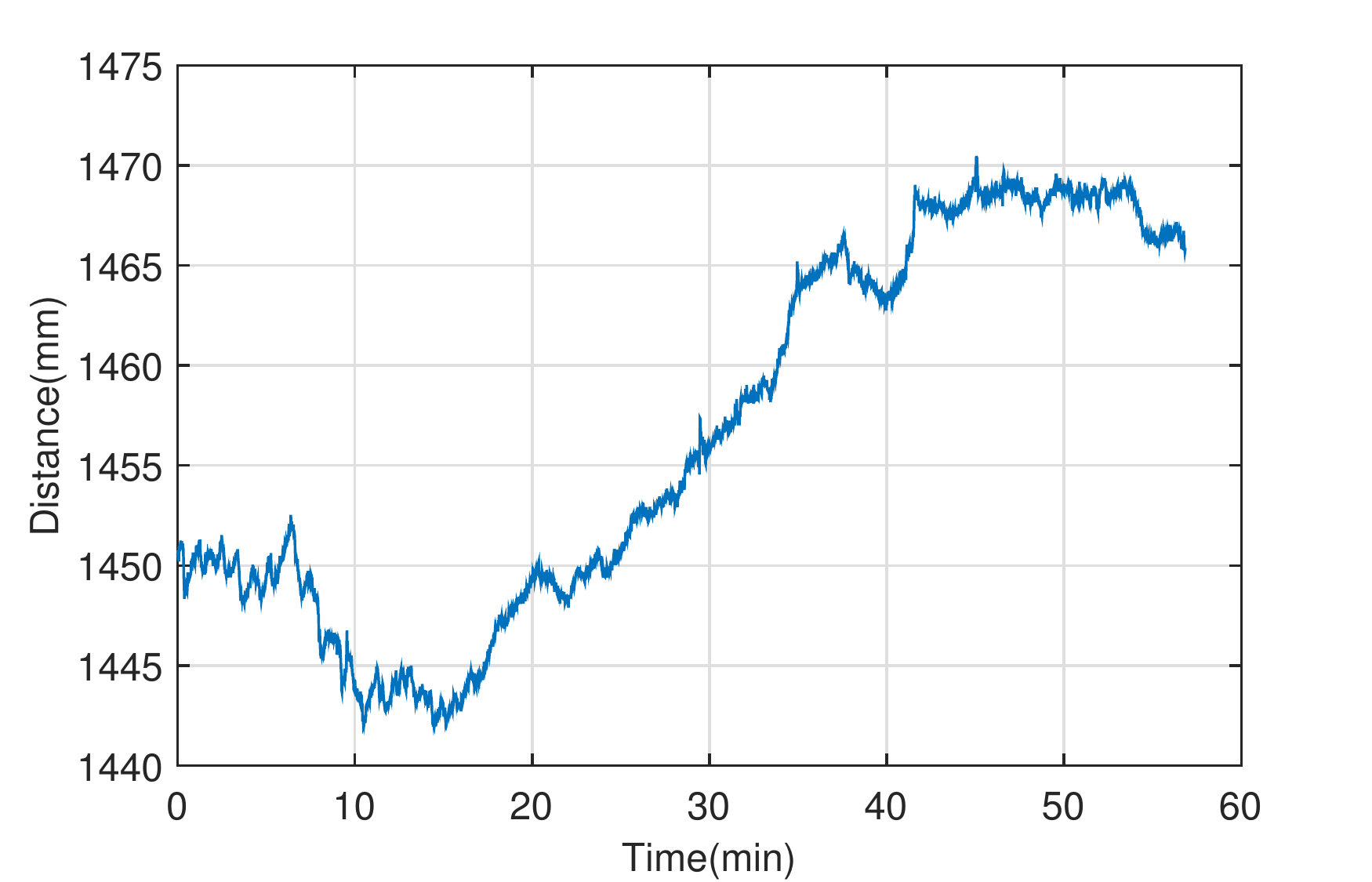}
%\DeclareGraphicsExtensions
\caption{Drift effect of RS-LiDAR}
\label{fig_drift}
\end{figure}

The distances to the target during the drift effects tests are presented in Fig.\ref{fig_drift}. It is apparent that there was a drop with more than $5$ mm of the measured distance over the first $15$ minutes. While after that, it began to rise until the $41$ minutes. This drift effect is probably due to the mirror deformation caused by the continuous increase of the operating temperature of the sensor. While unfortunately, the inside temperature module didn't open the data to users and we can only get the calibrated data from its official software. There appears not to be a significant correlation with range noise (i.e. standard deviation) because the noise level of the laser remained stable during the entire time of the tests. % As a result, the drift effect can be easily calibrated and the RS-LiDAR has stable performance after calibration. 

\subsection{Orientation Influence}
\begin{figure}
\centering
\includegraphics[width=3.3in]{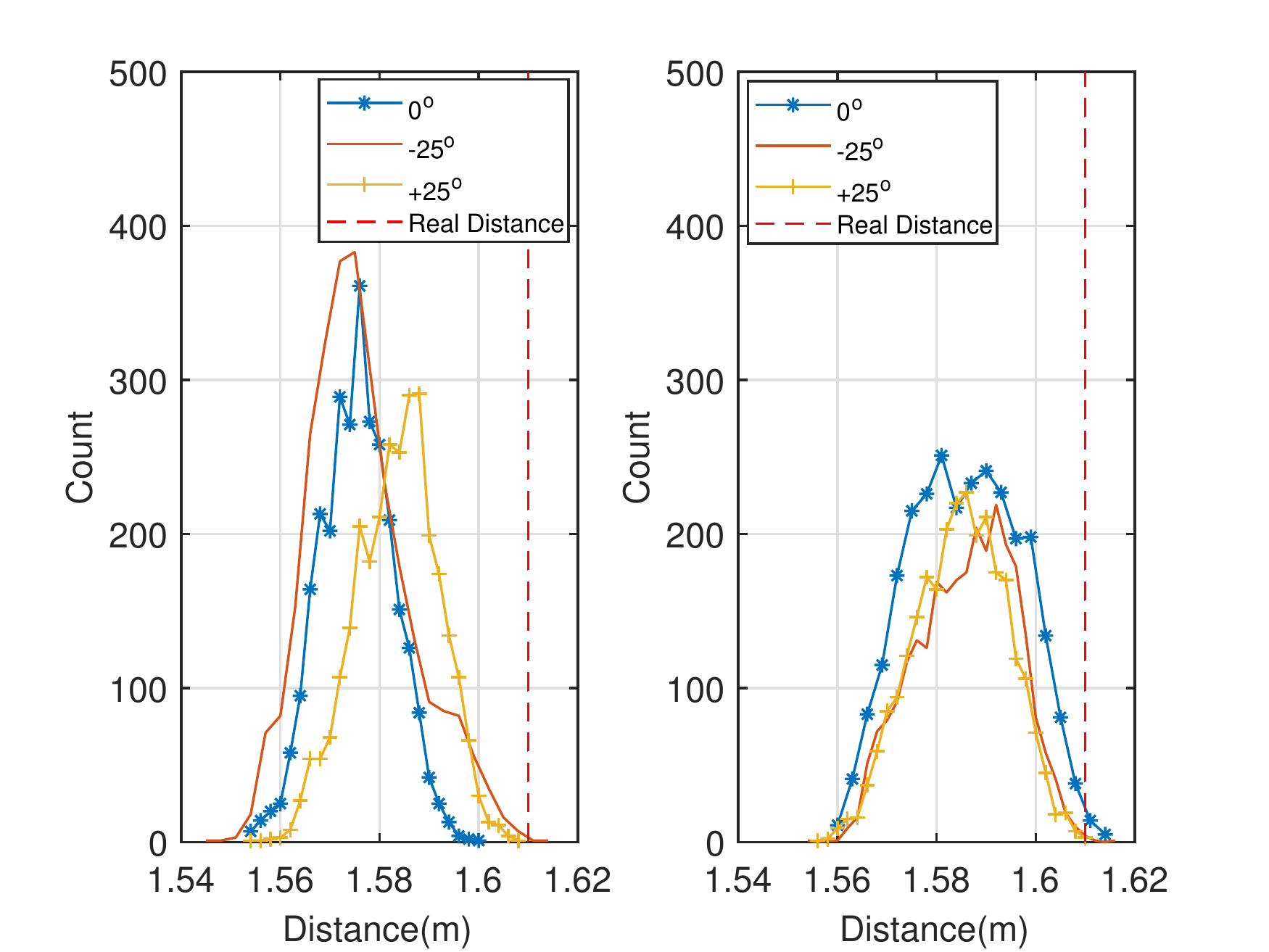}
%\DeclareGraphicsExtensions
\caption{Roll effect of RS-LiDAR (Left) and VLP-16 (Right) at 1.61 meters}
\label{rollrs}
\end{figure}

To determine whether the rotation is a significant influence on measurements, we settle the sensor to measure a white planary wall with different orientation. We controlled different roll angles with an index instrument, such as $0^\circ$, $-25^\circ$ and $+25^\circ$. The distance distributions are revealed in Fig.\ref{rollrs}. 

Fig.\ref{rollrs} presents that the influence of roll angles is negligible since the waves almost the same when sensors were in different roll angles. The mean distance values of $0^\circ$ for RS-LiDAR is 1576.4 mm, with 1584.5 mm at $25^\circ$ and 1577.5 mm at $-25^\circ$. The values for VLP-16 of $0^\circ$ is 1587.2 mm, with 1585.8 mm at $25^\circ$ and 1586.9 mm at $-25^\circ$. So the roll errors are less than 0.6\% within the roll angles from $-25^\circ$ to $25^\circ$. At the same time, in each line, the left parts and the right parts are almost symmetrical to each other, which also means that the two LiDARs show high accuracy in detecting the distances. When comparing the two LiDARs, the VLP-16 shows more independence on roll angles, while RS-LiDAR has less variance on the distances. It depicts that the average distance errors are similar, but the distribution of the VLP-16 is much consistent. % of the RS-LiDAR marginally outperforms VLP-16. 

%different but very close. slightly It's hard to distinguish the better one between two LiDARs.
%Furthermore, the autonomous cars would not have large orientation angle, so the data of RS-LiDAR will be rarely influenced by angles when it is used on a car.

\subsection{Performance of 16 Laser Beams}

\begin{figure}
\centering
\includegraphics[width=3.3in]{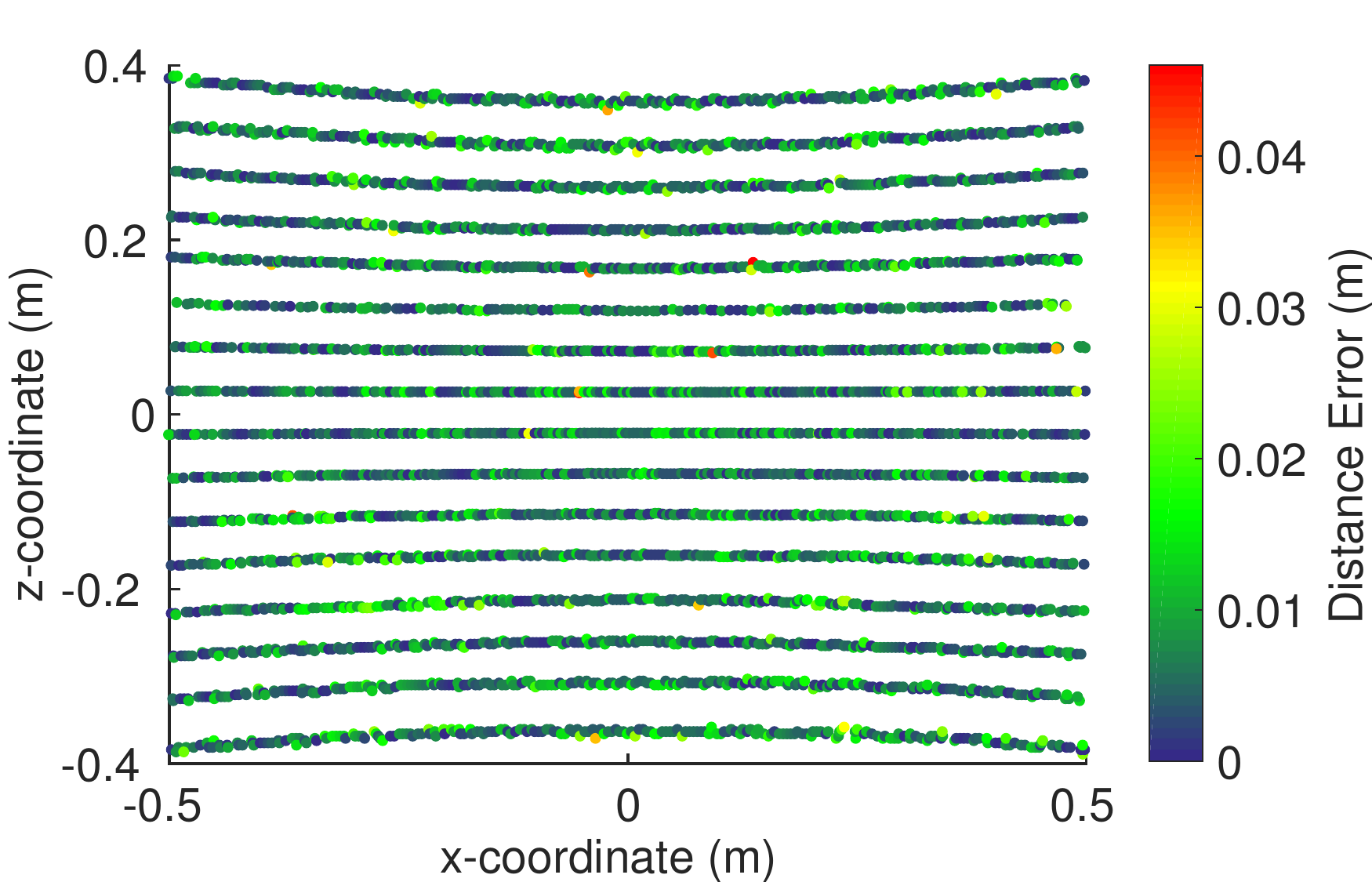}
\includegraphics[width=3.3in]{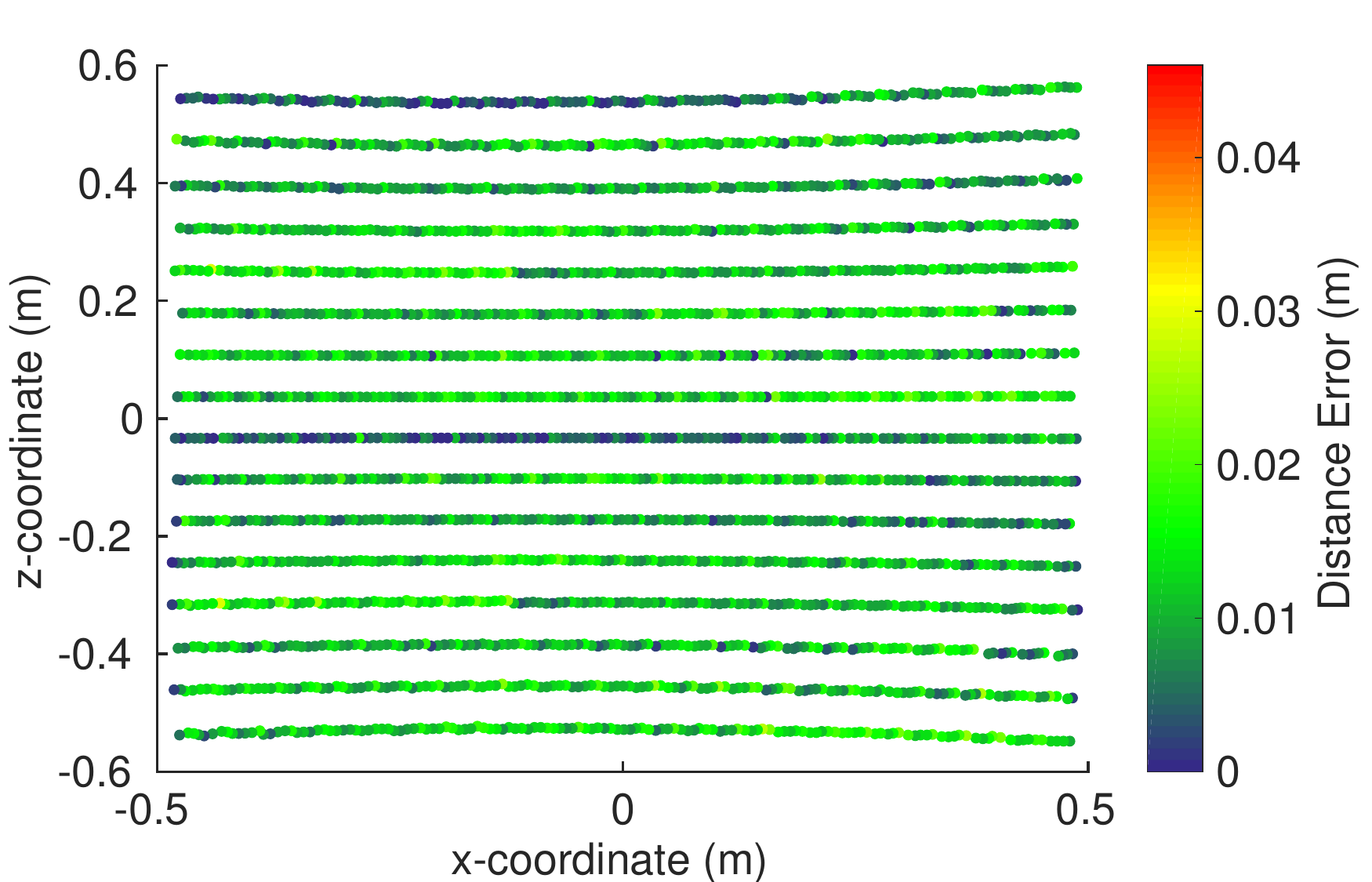}
\caption{Distance errors of \protect\\ RS-LiDAR (up) and VLP-16 (down)}
\label{diserrve}
\end{figure}

With the intention of figuring out whether there is a difference among the 16 laser beams of RS-LiDAR and VLP-16, we kept the sensors at 2.0 m far away from the target wall, fixing their central axis parallel to the wall and the y-axes vertical to the wall. Afterwards, we collected the detected data about the distances and computed each beam's standard deviation value of its y-axis values. The results are demonstrated in Fig.\ref{diserrve}. The 16 lines in each figure are data collected by their 16 laser beams. These two LiDARs show great accuracy in distance detection since the error rates are all about 0.2\% for each line. RS-LiDAR has better performance in this experiment because most of the distance errors of the data points for RS-LiDAR are less than 0.01 m while that for VLP-16 are between 0.01 m and 0.03 m. Also, the distance errors of the 16 lines for RS-LiDAR shows similar distribution but the top and center lines for VLP-16 are quite different with the others. In a nutshell, the 16 laser beams of RS-LiDAR have similar performances and outperforms the VLP-16.

We also did experiments on comparing the maximum detection distances of two LiDARs. The outcomes are in Table \ref{16beams}. It is apparent that the RS-LiDAR can detect much far than VLP-16. The max distances for the upper and lower laser beams may be a little bit smaller than the middle ones because of their angles, but the 16 beams still have similar detection ranges.

\begin{table}
\renewcommand{\arraystretch}{1.3}
\caption{Maximum detection distance of 16 laser beams}\label{16beams}
\scriptsize
\centering
\begin{tabular}{c|c|c|c|c}
\hline 
\multirow{3}{*}{\bfseries Beem No.} & \multicolumn{2}{c|}{\bfseries VLP-16} & \multicolumn{2}{c}{\bfseries RS-LiDAR} \\ 
\cline{2-5}
& \tabincell{c}{Effective\\ points} & \tabincell{c}{Max\\ Distance (m)} & \tabincell{c}{Effective\\ points} & \tabincell{c}{Max\\ Distance (m)} \\ 
\hline
-15$^\circ$ & 14 & 98.8 & 3 & 118 \\
-13$^\circ$ & 15 & 98.5 & 10 & 120 \\
-11$^\circ$ & 18 & 99.5 & 10 & 121 \\
-9$^\circ$ & 10 & 101 & 12 & 121 \\
-7$^\circ$ & 12 & 101 & 11 & 122 \\
-5$^\circ$ & 19 & 101 & 15 & 122 \\
-3$^\circ$ & 9 & 101 & 9 & 123 \\
-1$^\circ$ & 9 & 102 & 13 & 123 \\
1$^\circ$ & 24 & 103 & 18 & 123 \\
3$^\circ$ & 17 & 103 & 15 & 123 \\
5$^\circ$ & 10 & 101 & 16 & 123 \\
7$^\circ$ & 24 & 102 & 14 & 122 \\
9$^\circ$ & 8 & 101 & 10 & 122 \\
11$^\circ$ & 7 & 99.9 & 7 & 121 \\
13$^\circ$ & 19 & 99.5 & 15 & 120 \\
15$^\circ$ & 15 & 98.9 & 8 & 119 \\
\hline
\end{tabular}
\end{table}

\subsection{Effect of the Surface Color}
\label{surfacecolor}
\begin{figure}
\centering
\includegraphics[width=1.3in]{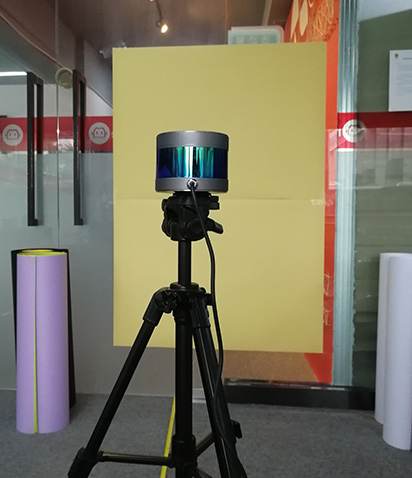}
%\DeclareGraphicsExtensions
\caption{The experiment setup on different colors}
\label{colortest}
\end{figure}
\begin{figure}
\centering
\includegraphics[width=3.3in]{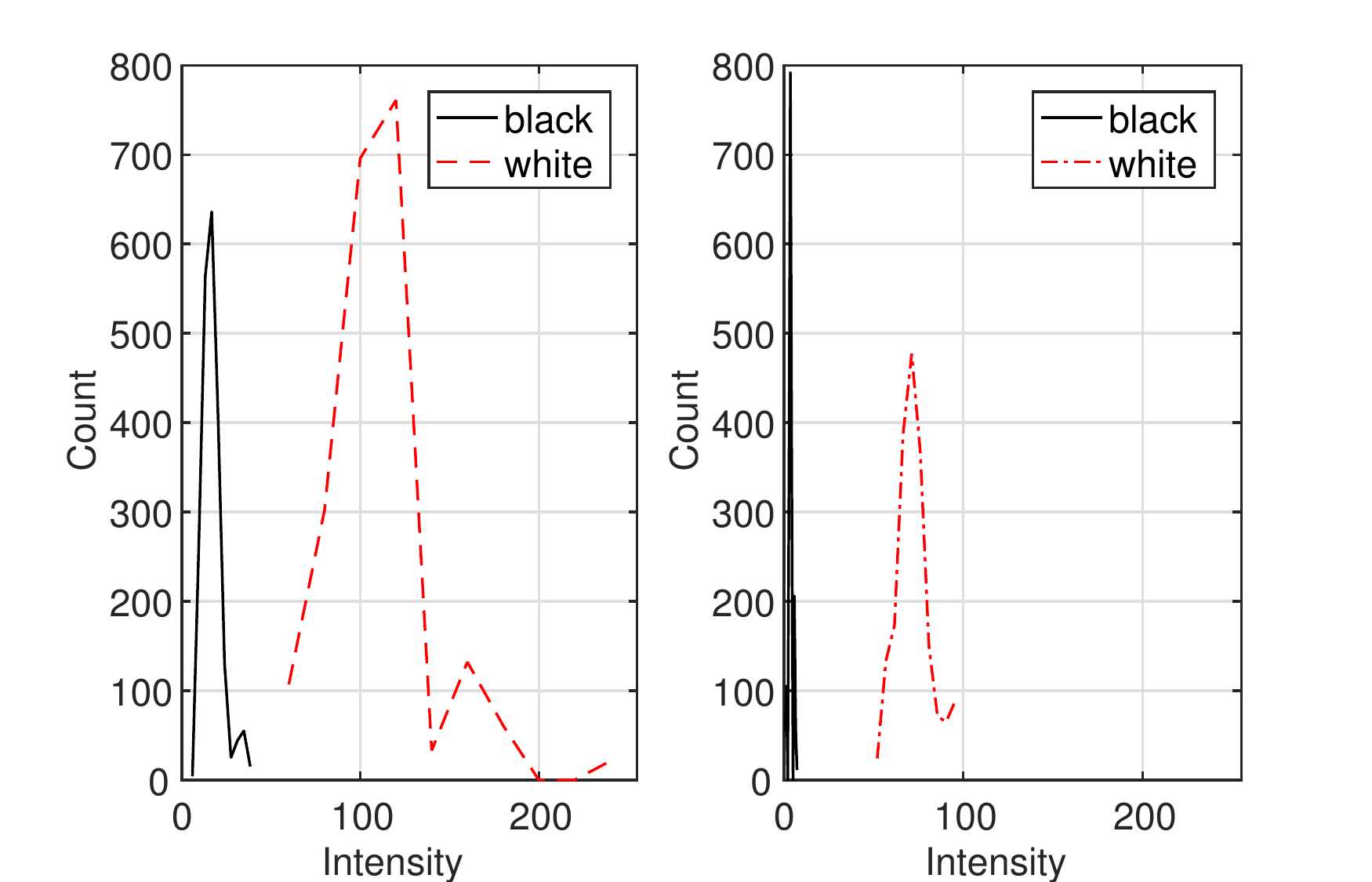}
\includegraphics[width=3.3in]{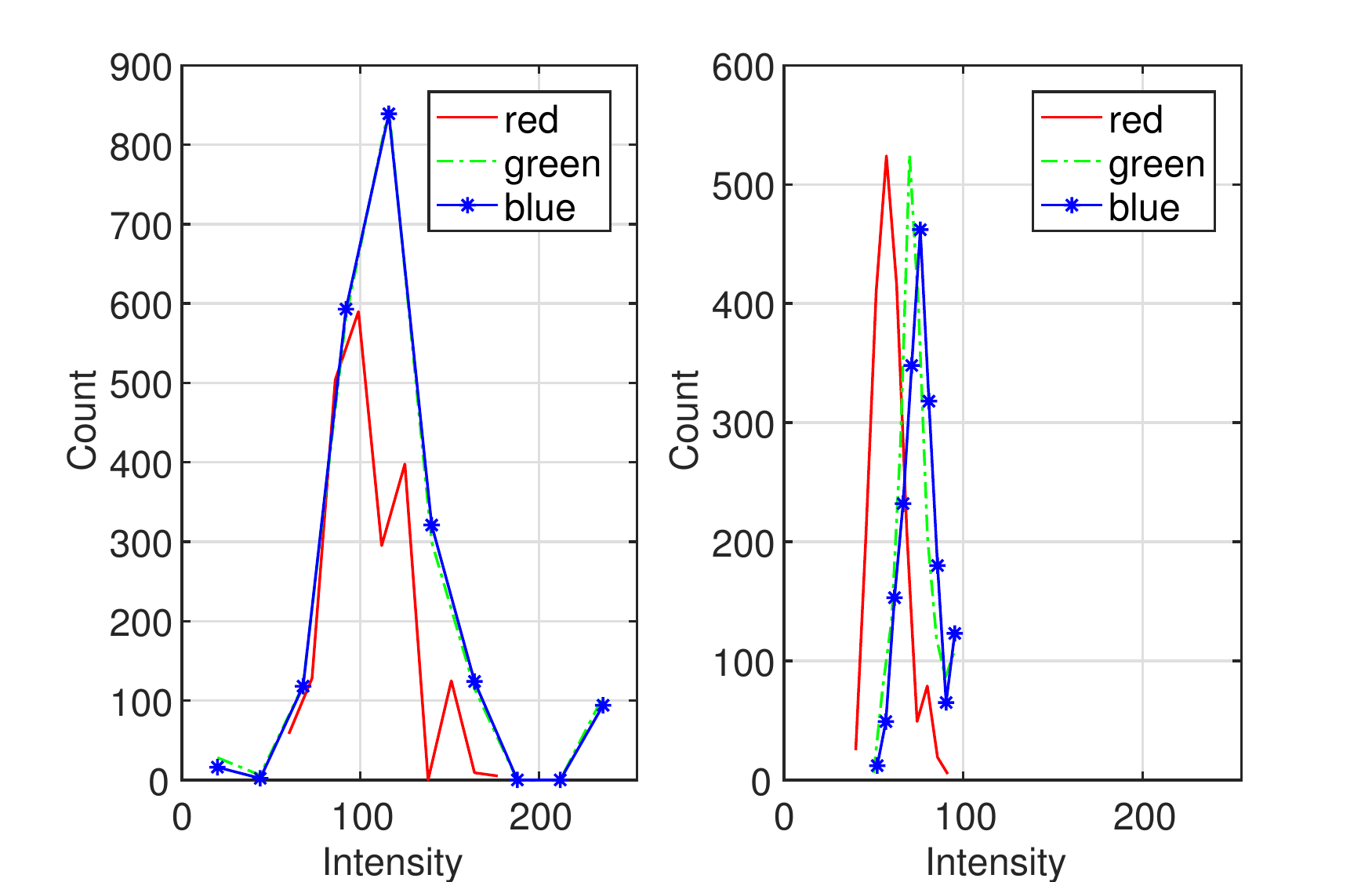}
\includegraphics[width=3.3in]{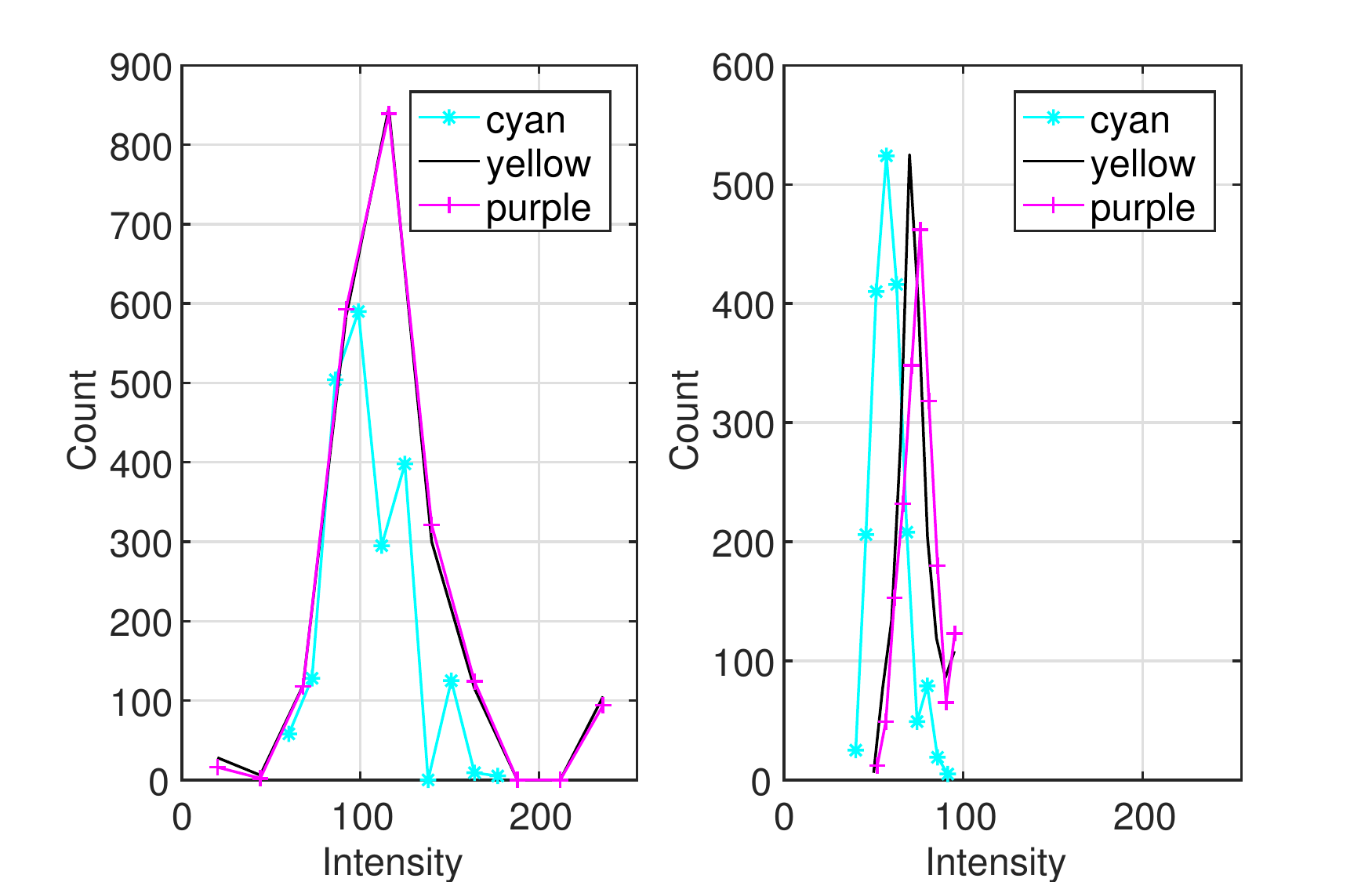}
\caption{Reflection intensity distributions of RS-LiDAR (Left) and VLP-16 (Right) for eight colors}
\label{fig_cyp}
\end{figure}
\begin{figure}
\centering
\includegraphics[width=3.3in]{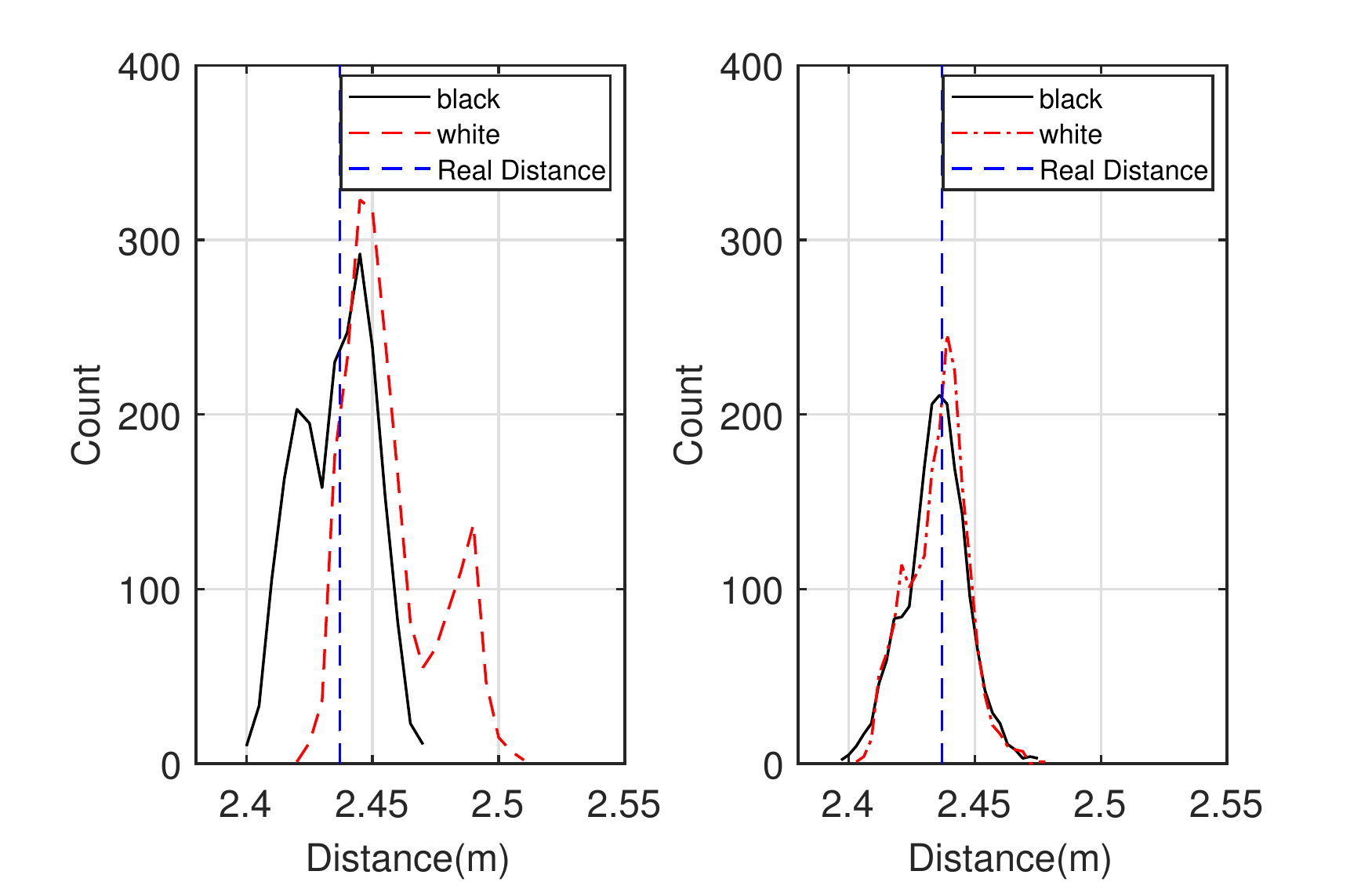}
\includegraphics[width=3.3in]{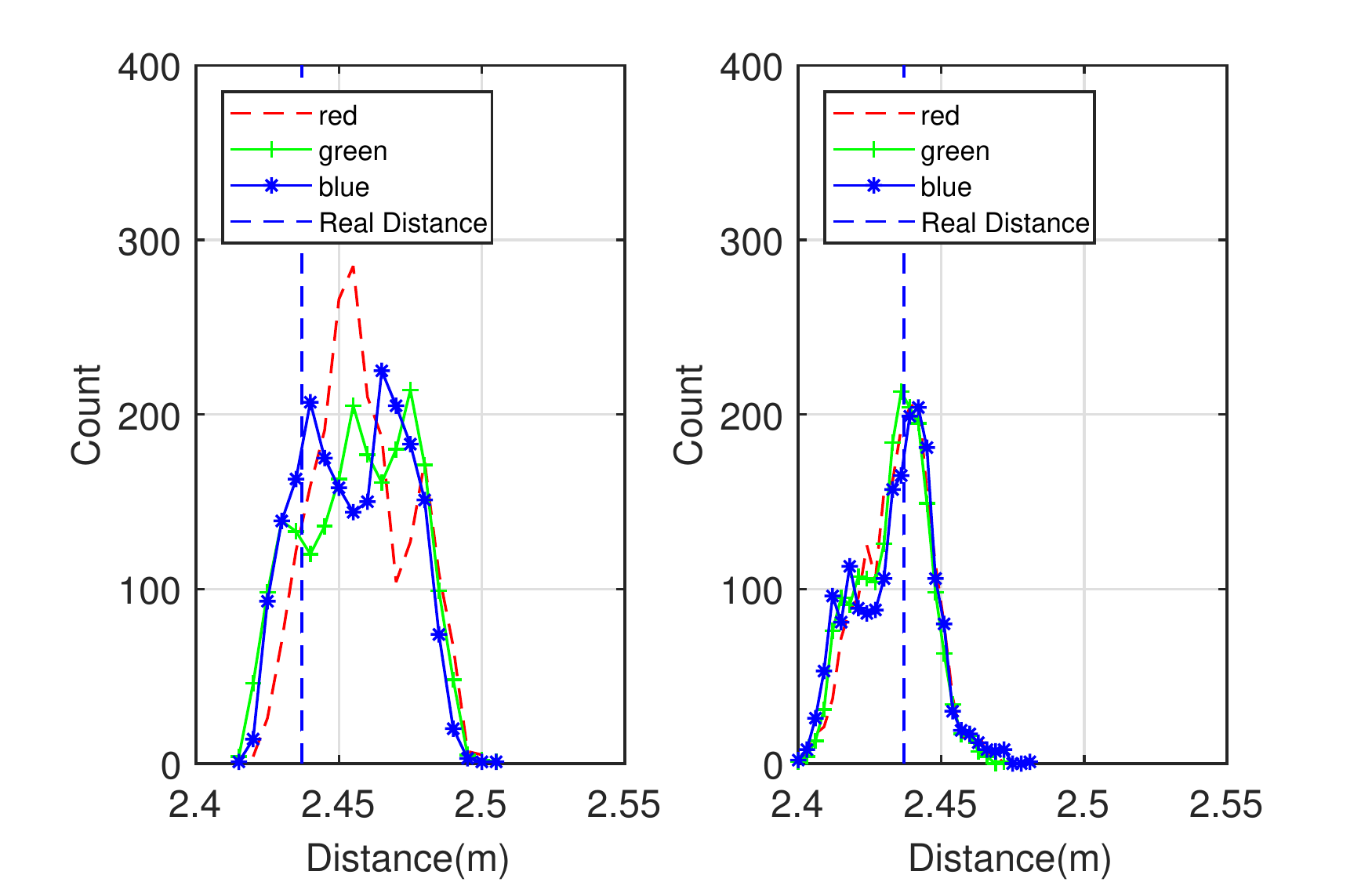}
\includegraphics[width=3.3in]{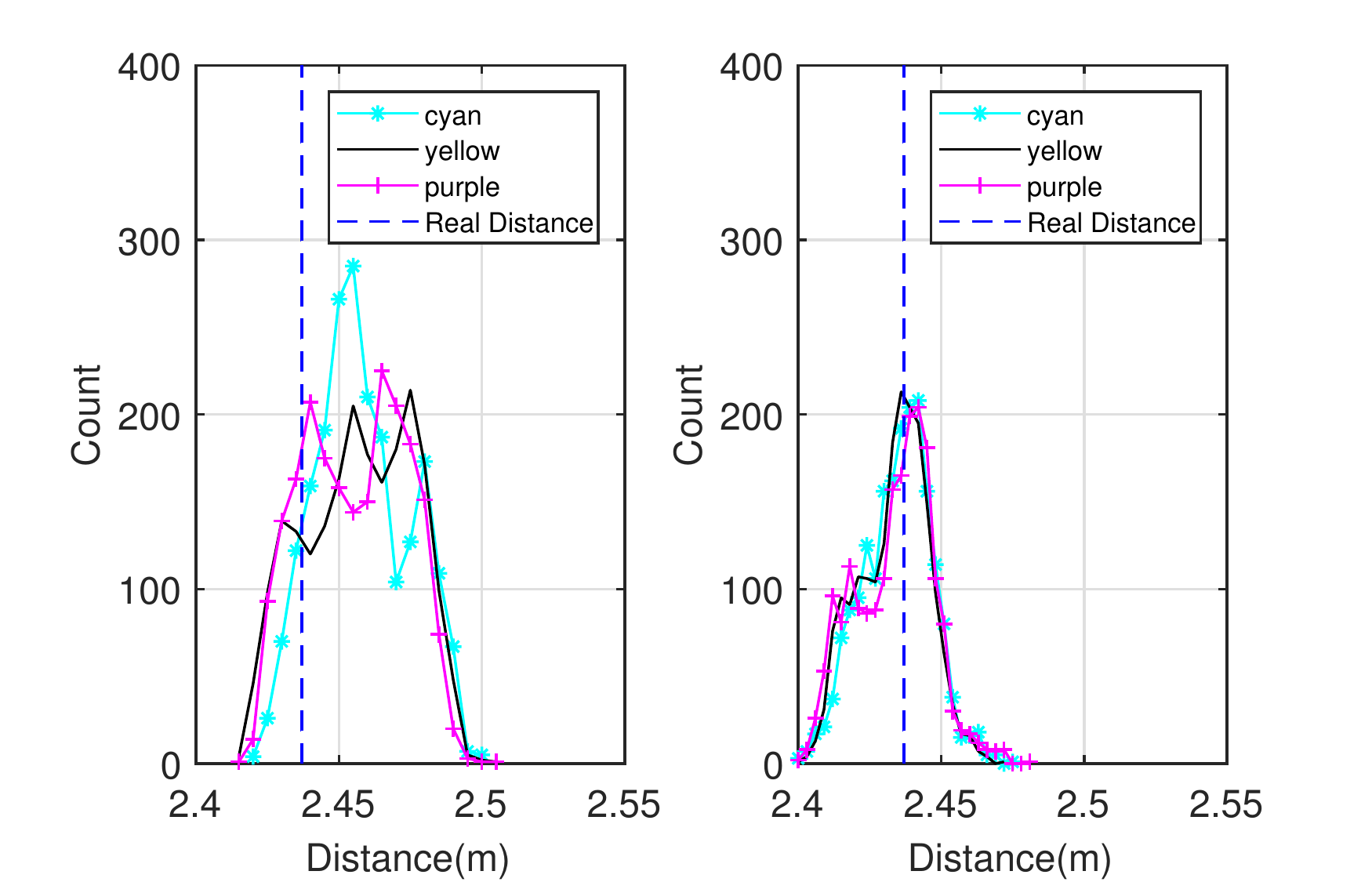}
\caption{Distances distributions of RS-LiDAR (Left) and VLP-16 (Right) at 2.437 meters}
\label{fig_cyp2}
\end{figure}

The colors and materials' reflectivity of the targets usually have effects on the reflection of lasers. RS-LiDAR and VLP-16 measure the reflectivity of an object with a 256-bit resolution which is independent of laser power. In order to get the LiDARs' performances when detecting different colors of the plane surface target, the three primary colors in color wheel which are red, green, blue and three secondary colors which are cyan, yellow, purple were tested. Besides, two different gray levels of colors, white and black, were also tested. Those eight colors targets are all the same materials of papers and were all posted on right the same place during each test. Fig.\ref{colortest} shows the experiment environment. The reflection intensity distributions are shown in Fig.\ref{fig_cyp} and the distance distributions in Fig.\ref{fig_cyp2}. 
 
According to Fig.\ref{fig_cyp}, for both of the LiDARs, the intensity of black surface is lower if compared with the white one. The other six colors show the similar graphs in intensity distributions. As a result, there is little differences in the reflection intensity of the targets with a majority of the common colors while the dark objects would be a great influence on the reflection intensity. RS-LiDAR got more precious reflection intensities from the black surface than VLP-16. This illustrates that RS-LiDAR is more sensitive, which is also revealed by the process of map construction in Fig.\ref{fig_rslidarmap} and \ref{fig_velodynemap}. RS-LiDAR has a broader range of intensity which helps to show more details of the targets.

On the basis of Fig.\ref{fig_cyp2}, it illustrates that the differences of measured distances to the targets with different colors are not very obvious but VLP-16 has less variance in the distance distributions. The mean distances measured by RS-LiDAR are 2437.9 mm, 2459.4 mm, 2460.7 mm, 2459.4 mm, 2458.0 mm, 2457.1 mm, 2458.9 mm and 2458.2 mm for the color black, white, red, green, blue, cyan, yellow and purple respectively, with the maximum error 22.8 mm (0.9\%). The data for VLP-16 are 2436.1 mm, 2436.8 mm, 2436.1 mm, 2434.9 mm, 2435.2 mm, 2435.7 mm, 2435.5 mm and 2436.5 mm, with the maximum error 0.1 mm (0.003\%). The distances diversity with colors of two LiDARs are all acceptable, but the distance measured by RS-LiDAR for a black target is less than the others. In conclusion, if extremely precision on black targets is not needed, for example, on autonomous cars, colors of the targets won't make much impact on the measured distances and RS-LiDAR can be used as well as VLP-16.

%\begin{figure}
%\centering
%\includegraphics[width=3.3in]{colorDis}
%%\DeclareGraphicsExtensions
%\caption{Distances Measured by RS-LiDAR and VLP-16 for eight colors}
%\label{fig_disRS}
%\end{figure}

\subsection{Dependences on Target Distance}
\begin{figure}
\centering
\includegraphics[width=3.3in]{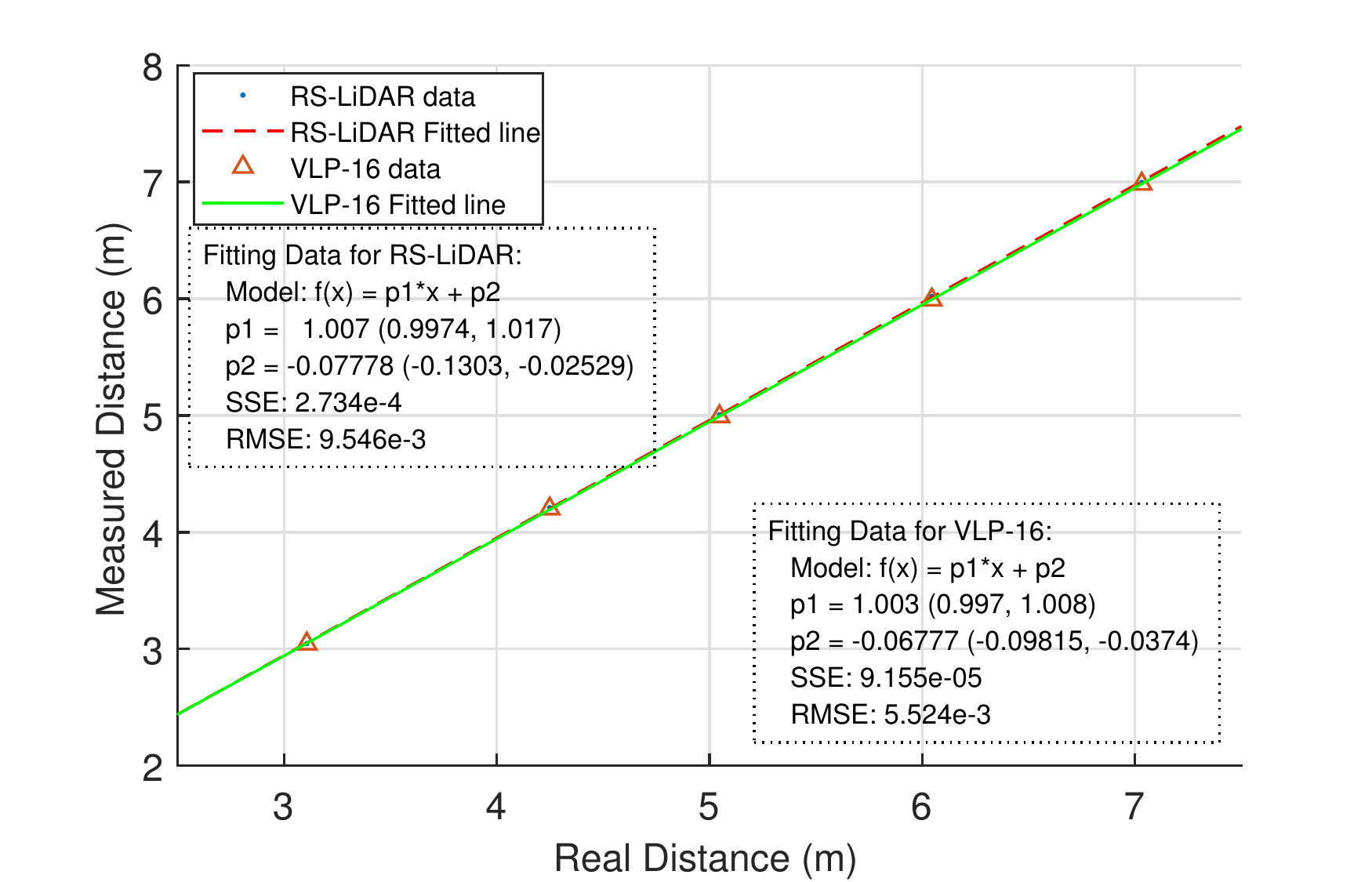}
%\DeclareGraphicsExtensions
\caption{Comparison of the distances measured by LiDARs and truth}
\label{fig_mean}
\end{figure}

Aiming to get an idea of the dependencies on the target distance, a measurement to the same surface with different distances has been performed. Five distances between the target wall and the LiDAR, which is in every $1$ m from $3$ m to $7$ m, were measured. The measured distances are shown in Fig.\ref{fig_mean}, taking the real distance from a laser range finder as a reference. By adding a fitting on the data for each LiDAR, it demonstrates that the fitted lines for two LiDARs from 3 m to 7 m are linear ones which mean the distances these two LiDARs measured are in great precision. When applying the fitting, we also computed the SSE and RMSE with the equations (\ref{equa_sse}) and (\ref{equa_rmse}).
\begin{equation}
\label{equa_sse}
SSE=\sum^{n}_{i=1}w_i(y_i-\hat{y}_i)^2
\end{equation}
\begin{equation}
\label{equa_rmse}
RMSE=\sqrt{\frac{1}{n}\sum^{n}_{i=1}w_i(y_i-\hat{y}_i)^2}
\end{equation}

The SSE and RMSE for RS-LiDAR and VLP-16 showing in Fig.\ref{fig_mean} are all in tiny orders of magnitude. The SSE value of RS-LiDAR is 2.734e-4 meter and that of VLP-16 is 9.155e-05 meter. The RMSE value of RS-LiDAR and VLP-16 are 9.546e-3 meter and 5.524e-3 meter. This is another proof of the great precision of two LiDARs, though the VLP-16 had a better performance. 

%The reflection intensity data shows a surprising trend in Fig.\ref{fig_intens}. Normally, the reflection intensity should be less when the sensor went farther to the target because of the atmospheric absorption to lasers. But in this experiment, the result was totally on the contrary. With the distances increasing, the intensity of the captured reflection went along with an obvious increase. That phenomenon may caused by the non-parallelization of each laser beam. It is nearly impossible to make sure that the laser beam ejected is a parallel straight light shoot. In RS-LiDAR, the laser beam is a convergent light shoot whose focal point is farther than 8 meters. So in the experiment, if the LiDAR is put near to the target surface, the light spot on the surface is a little bit larger so that more light is dispersed with diffuse reflection and less light is reflected back to the sensor. Therefore, the reflection intensity may be larger when the sensor were farther.

\subsection{Common Scenes and Objects Detection}
\begin{figure}%[!t]
\centering
\subfloat[Car]{\includegraphics[height=0.14\textwidth]{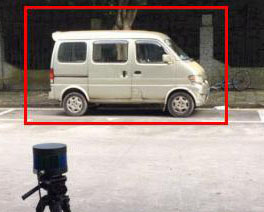}%
\label{fig_car}}
\hfil
\subfloat[Person]{\includegraphics[height=0.14\textwidth]{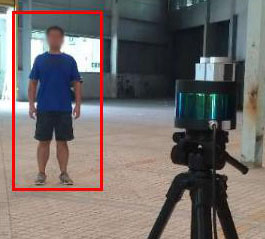}%
\label{fig_person}}
\hfil
\subfloat[Telegraph Pole]{\includegraphics[height=0.14\textwidth]{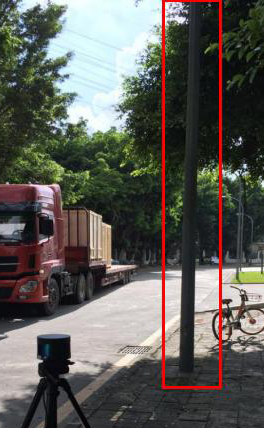}%
\label{fig_pole}}
\hfil
\subfloat[Car 3m]{\includegraphics[height=0.14\textwidth]{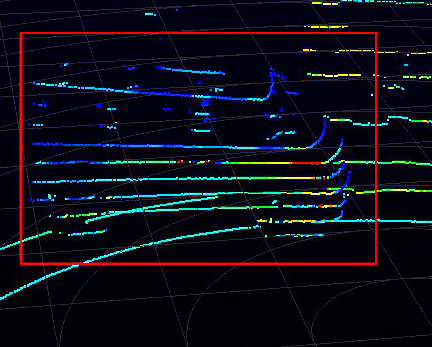}%
\label{fig_first_case}}
\hfil
\subfloat[Person 5m]{\includegraphics[height=0.14\textwidth]{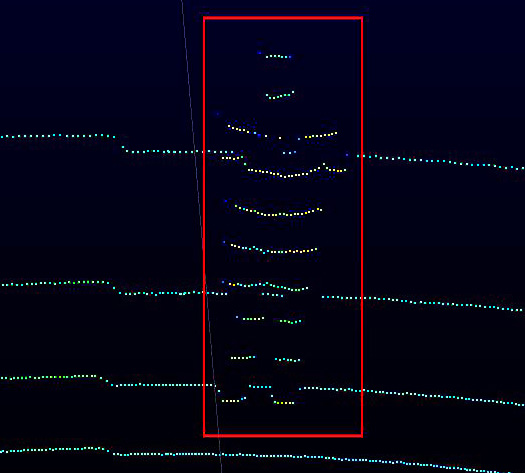}%
\label{fig_4_case}}
\hfil
\subfloat[Telegraph Pole 5m]{\includegraphics[height=0.14\textwidth]{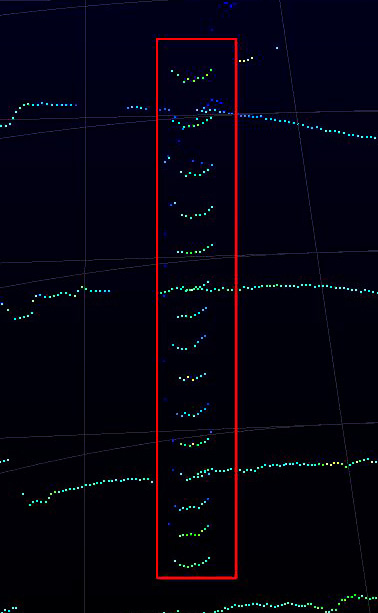}%
\label{fig_7_case}}
\hfil
\subfloat[Car 3m]{\includegraphics[height=0.14\textwidth]{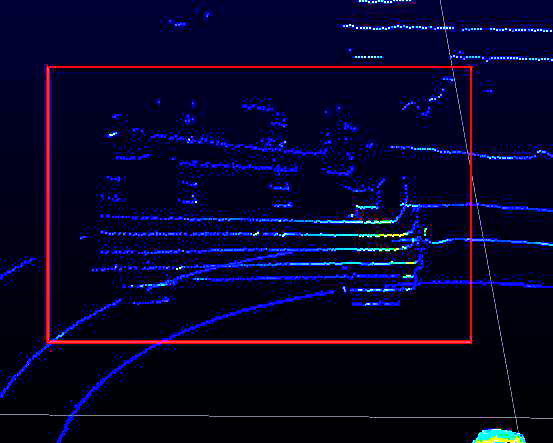}%
\label{fig_10_case}}
\hfil
\subfloat[Person 5m]{\includegraphics[height=0.14\textwidth]{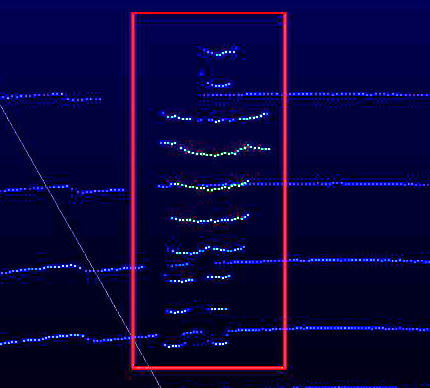}%
\label{fig_13_case}}
\hfil
\subfloat[Telegraph Pole 5m]{\includegraphics[height=0.14\textwidth]{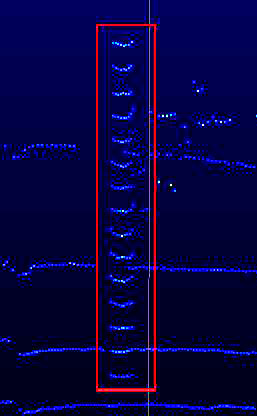}%
\label{fig_16_case}}
\caption{Common scenes detection of RS-liDAR (middle) and VLP-16 (down)}
\label{fig_common_scene}
\end{figure}

In Fig.\ref{fig_common_scene}, the nine pictures show the data points from RS-LiDAR and VLP-16 when detecting a series of objects which are common in life. The data points from two LiDARs seems really similar in these pictures. With these pictures, readers can also have a comparison between these two LiDARs and consider whether 16 beams LiDAR is suitable for the work to be done.

\section{Investigation of Calibration Nethods}%-------------------------------------------------4444444444444444444444444444444
\label{sec_calibri}
There are many papers which introduce the methods of laser scanner calibration, which could be expanded to single-beam LiDAR calibration and multi-beam LiDAR calibration. Such as papers from Jesse Levinson and Sebastian Thrun\cite{Levinson}, Mark Sheehan and Alastair Harrison\cite{Sheehan}, C.L. Glennie and A.Kusari\cite{Glennie}, Teichman \cite{Teichman} and Muhammad\cite{Muhammad}. In order to achieve an accurate calibration, their papers describe the detailed works of a specific calibration method for lasers. In the current paper, we would simply have a talk about the calibration method.

This section tries to analyze the linearity and accuracy of the measurement results and establish a calibration model for RS-LiDAR. We chose 3 colors and segmented out about 4000 measurement values each time. Since there were not many kinds of colors analyzed, this evaluation cannot claim to provide an extended and very precise calibration model, but it can still give an idea of the effect of the errors in the function of these parameters.

%\begin{table}
%\renewcommand{\arraystretch}{1.3}
%\caption{Relative Errors of Different Colors}
%\label{relative_error}
%\scriptsize
%\centering
%\begin{tabular}{l|c}
%\hline
%\bfseries Color &\bfseries Relative Error (mm) \\
%\hline
%red & -5.6318 \\
%green & -2.0359 \\
%blue & 3.5143 \\
%cyan & 1.1740 \\
%yello & -5.4466 \\
%purple & -7.4404 \\
%\hline
%\end{tabular}
%\end{table}

To get a calibration model, tests with different distances are needed to get different relative errors or absolute errors. Thus, a correction would be calculated with those errors. To get correction parameters of sensor's orientation, we should calculate the errors with different orientation angles so that we will know how to compensate and get the nearest right measurement. Finally, with a gyroscope on the sensor, we can know the real-time correction.

\section{Conclusion}%-------------------------------------------------5555555555555555555555555555555
This paper presented the characterization of the RS-LiDAR, in comparison with VLP-16. We analyzed several performances such as temperature drift, orientation influence, differences in 16 laser beams, surface color influence and distance influence. The results of the measurements showed that the characterizations of RS-LiDAR are equivalent to the VLP-16 in key characteristics.

\begin{figure}
\centering
\includegraphics[width=3.3in]{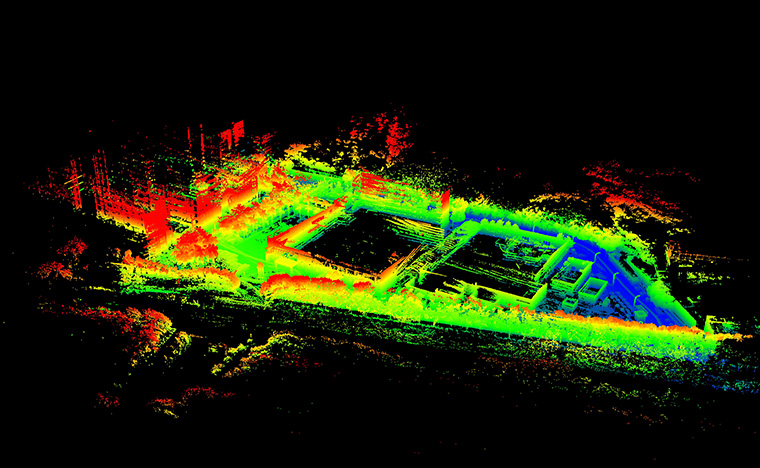}
%\DeclareGraphicsExtensions
\caption{The map built by RS-LiDAR}
\label{fig_rslidarmap}
\end{figure}

\begin{figure}
\centering
\includegraphics[width=3.3in]{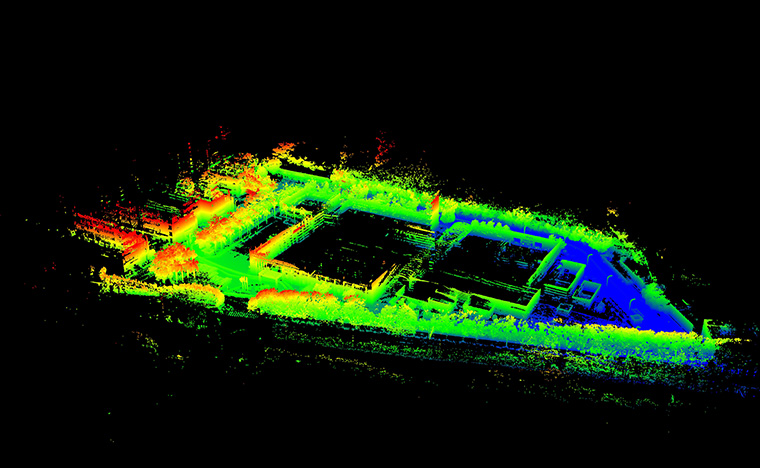}
%\DeclareGraphicsExtensions
\caption{The map built by VLP-16 with the same path as that in Fig.\ref{fig_rslidarmap}}
\label{fig_velodynemap}
\end{figure}

We also tested the application in 3D perception. A golf-car mounted with both sensors travels along a loop of our campus. 3D mapping was run based on our recent modular implementation\footnote{https://ram-lab.com/research}. The identical algorithms with the same parameters were adopted across the experiments. The mapping results were shown in Fig.\ref{fig_rslidarmap} and Fig.\ref{fig_velodynemap}, built by RS-LiDAR and VLP-16, respectively. The former has a more widely scene but a little bit less distinct noisy points when compared with the latter. As a result, RS-LiDAR has better performance in 3D mapping. The validation in Section \ref{surfacecolor} on the sensitivity to surface characteristic and distances revealed the reasons.
%\label{sec_experi}

In all, RS-LiDAR is proved entirely qualified to replace VLP-16, thanks to its sufficient supplies, lower price, and similar performance. The 32-beam LiDAR from Robosense was also newly released. We will test on the new sensor in the near future.


\section*{Acknowledgment}%-------------------------------------------------6666666666666666666666666666666666666666666
This paper is supported by the Research Grant Council of Hong Kong SAR Government, China, under project No. 16206014 and No. 16212815; National Natural Science Foundation of China No. 6140021318, awarded to Prof. Ming Liu; National Natural Science Foundation of China No. 61603376, awarded to Dr. Lujia Wang.

\end{document}